\documentclass{article} 
\usepackage{iclr2015,times}

\usepackage{fixltx2e} 		

\usepackage{algorithm} 		
\usepackage{algorithmicx} 	
\usepackage{algpseudocode} 	

\usepackage{todonotes}		

\usepackage{amsmath}		

\usepackage{amsfonts}		
\usepackage{bm} 			  
\usepackage{mathtools} 		

\usepackage{microtype}		

\usepackage{graphicx}		
\usepackage{wrapfig}
\usepackage{float} 			
\usepackage{caption}			
\usepackage{subcaption}		

\usepackage{tabularx}  		
\usepackage{multirow}		

\usepackage{enumerate}
\usepackage[ampersand]{easylist} 

\usepackage{hyperref}		
\usepackage{url}

\usepackage{pdflscape} 		

\newcommand{\thetav}{\vec{\theta}}

\DeclareMathOperator*{\argmax}{arg\,max}

\newcommand{\params}{\thetav}
\newcommand{\lr}{\alpha} 	

\newcommand{\momlr}{\mu}
\newcommand{\batch}{B}
\newcommand{\Alpha}{A}
\newcommand{\data}{X}
\newcommand{\alphas}{\vec{\alpha}}

\title{Hot Swapping for Online Adaptation of \\Optimization Hyperparameters}

\author{
Kevin M. Bache \& Padhraic Smyth\\
Department of Computer Science\\
University of California, Irvine\\
Irvine, CA 92697, USA \\
\texttt{\{kbache,smyth\}@ics.uci.edu} \\
\And
Dennis DeCoste \\
Machine Learning Group\\
eBay Research Labs\\
San Jose, CA 95125, USA \\
\texttt{ddecoste@ebay.com} \\
}

\graphicspath{{./figs/}}
     
\iclrfinalcopy 

\begin{document}

\maketitle

\begin{abstract}
We describe a general framework for online adaptation of optimization hyperparameters by `hot swapping' their values during learning. We investigate this approach in the context of adaptive learning rate selection using an explore-exploit strategy from the multi-armed bandit literature. Experiments on a benchmark neural network show that the hot swapping approach leads to consistently better solutions compared to well-known alternatives such as AdaDelta and stochastic gradient with exhaustive hyperparameter search.
\end{abstract}

\section{Introduction}
\label{sec:intro}
In this paper, we introduce a new stochastic gradient method with adaptive learning rate selection based on the insight that optimization hyperparameters may be freely `hot swapped' in the middle of the learning process\footnote{The notion of model-based hot swapping of algorithms or their parameters has been previously considered the study of the `dynamic algorithm selection' or `dynamic algorithm configuration' problems \citep{gagliolo2006learning}, of which learning rate selection may be considered a specific instantiation.  Most past work in this area has been focused on combinatorial optimization problems, where here we consider a continuous optimization problem.}.

Where existing adaptive learning rate algorithms are based on running curvature estimates of the local loss surface \citep{schaul_no_2012,zeiler_adadelta:_2012}, we present a procedure which recasts learning rate selection as an explore-exploit problem which can be addressed using existing solutions to multi-armed bandit problems.  This method is straightforward to implement, retains the runtime characteristics and memory footprint of stochastic gradient, and outperforms existing methods on a common benchmark task. 

\section{Algorithm}
The basis of the proposed algorithm is the observation that optimization hyperparameters such as learning rate and momentum may be freely `hot swapped' during optimization runs.  This is in contrast to model hyperparameters, such as hidden layer size or unit type, which cannot be changed so easily during learning.  This approach can also be contrasted to traditional hyperparameter search strategies such as grid search, random search, or Bayesian optimization which set optimization hyperparameters in an outer loop and treat learning as an inner loop \citep{bergstra_random_2012,snoek_practical_2012}.  

Instead, we propose to observe the optimization process under a variety of hyperparameter settings and to preferentially continue to use those settings which have performed best in the past.  We do this by maintaining a meta-model of hyperparameter performance.  The general hot swapped optimization procedure is defined in algorithm \ref{alg:hot_swap}.  

Many meta-models may work well for this task, but in this work we cast the problem of learning rate selection into an explore-exploit framework and choose a discounted upper confidence bound (DUCB) model for hyper-parameter selection.  In brief, we seek an algorithm that `explores' the space of possible learning rates---in order to learn which ones perform best on the given problem---while reserving most of its time to `exploit' the best performing rates---by repeatedly using them to update model parameters.  The upper confidence bound algorithm is a common choice for tackling explore-exploit problems, and its discounted form achieves the optimal regret bound up to a logaritmic factor for rapidly shifting reward distributions \citep{garivier_upper-confidence_2008}.

The procedure for hot swapped optimization with a DUCB model is listed in algorithm \ref{alg:hot_swap_ducb} with the full details given in algorithm \ref{alg:ducb_helpers}. We assume that we have a finite set of learning rates to select from, $\alphas \equiv \{\alpha_1, ..., \alpha_K\}$, a postive objective function to be minimized $f(\params; \batch)$, along with its gradient $g(\params; \batch)$, both of which can be evaluated at a point in parameter space $\params$ for a given data batch $\batch$.  

We define the `reward' granted to a given learning rate $\alpha_k$ as $r = \log(f(\params_0; \batch)) - \log(f(\params_k; \batch))$, where $\params_0$ represents the (non-hyper) parameters at the beggining of the current iteration and $\params_k \equiv \params_0 - \alpha_k g(\params_0, \batch)$ represents the parameters obtained by choosing learing rate $\alpha_k$.  We use a logarithmic scaling of the rewards which treats multiplicative reductions of the objective function $f$ as equally valuable, a useful feature given the exponential slowdown of optimization progress which is often observed in practice.  The $\alpha$ value chosen at each step is selected by the DUCB algorithm in the usual way (see the function \textsc{GetDucbSuggestedIndex} in algorithm \ref{alg:ducb_helpers} for details).

The DUCB model will periodically seek to explore different learning rates as the optimization procedure progresses.  This introduces the potential to take catastrophically large steps which could discard progress made up to the current time.  To prevent this, we perform a line search across learning rates to find the best learning rate value for each minibatch.  The line search starts from the learning rate proposed by the DUCB algorithm and decreases through the other available learning rates until it finds one which lowers the current minibatch's objective function value below the value it held before the current update.  Because the line search is only performed on the current minibatch of data, it still takes linear time in batch size and problem dimension, just like simple SGD\footnote{Minibatch line search is not without precedent \citep{ngiam_optimization_2011,roux_stochastic_2012}, though it has received little direct attention in the past}.

In practice, these two processes work well together. The line search prevents the bandit algorithm's tendency to explore catastrophically large step sizes, while the bandit algorithm's carefully chosen initial step sizes reduce the number of minibatch objective function evaluations from the large number required by a vanilla minibatch line search to a much smaller quantity.

\section{Initial Results}
\label{sec:results}
We test the efficacy of this procedure\footnote{This experiment was conducted using Theano, PyLearn2, and a cluster of computers with NVIDIA GRID K520 GPUs} on a neural network based on the MNIST dataset.  MNIST is comprised of 60,000 28x28 pixel black and white images of handwritten digits.  The task is to classify each image as a number `0' through `9'.  

We show results from fully connected feed-forward network with 500 sigmoidal units in the first hidden layer, 300 sigmoidal hidden units in the second hidden layer, and a final 10-way softmax output.  We trained on the first 50,000 images of the training set. 

We perform an exhaustive search of SGD hyperparameters for comparison.  We consider all combinations of the following parameters: initial learning rates in $\lr_0 \in \{1.0, 0.3, 0.1, 0.03, 0.01, 0.003\}$, per-epoch learning rate multipliers of $\eta \in \{0.99, 0.995, 1.0\}$, momentum coefficients $\momlr \in \{0.0, 0.5, 0.7, 0.9\}$ and batch sizes $\in \{64, 128, 256, 512, 1024\}$ for a total of 360 SGD settings.  

We also compare against AdaDelta, another widely used adaptive learning rate algorithm, using the hyperparameters described in \cite{zeiler_adadelta:_2012} for MNIST: $\epsilon = 10^{-6}$ and decay rate factor of $0.95$ across batch sizes $\in \{64, 128, 256, 512, 1024\}$.

Finally, we tested hot swapped optimization with a DUCB step size model and batch sizes $\in \{64, 128, 256, 512, 1024\}$\footnote{It's worth noting that batch size is the one significant hyperparameter for the hot swapping DUCB algorithm.  It is similar in this regard to other adaptive learning rate methods such as AdaDelta and No More Pesky Learning Rates.}.

\begin{figure}[ht]
        \centering
        \begin{subfigure}{.5\textwidth}
                \includegraphics[width=\textwidth]{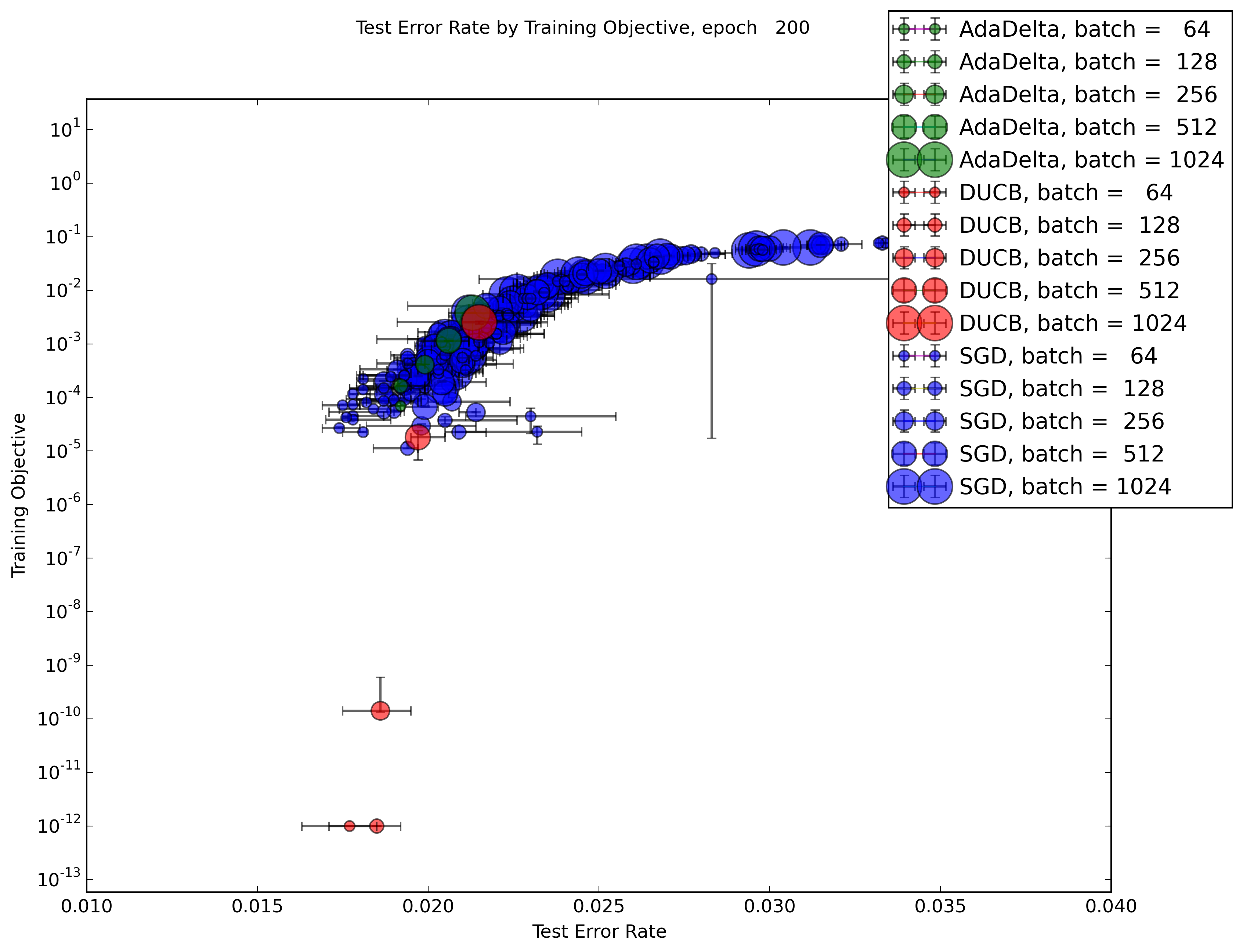}
                \caption{Epoch 200}
                \label{fig:500-300_scatter200}
        \end{subfigure}%
        \begin{subfigure}{.5\textwidth}
                \includegraphics[width=\textwidth]{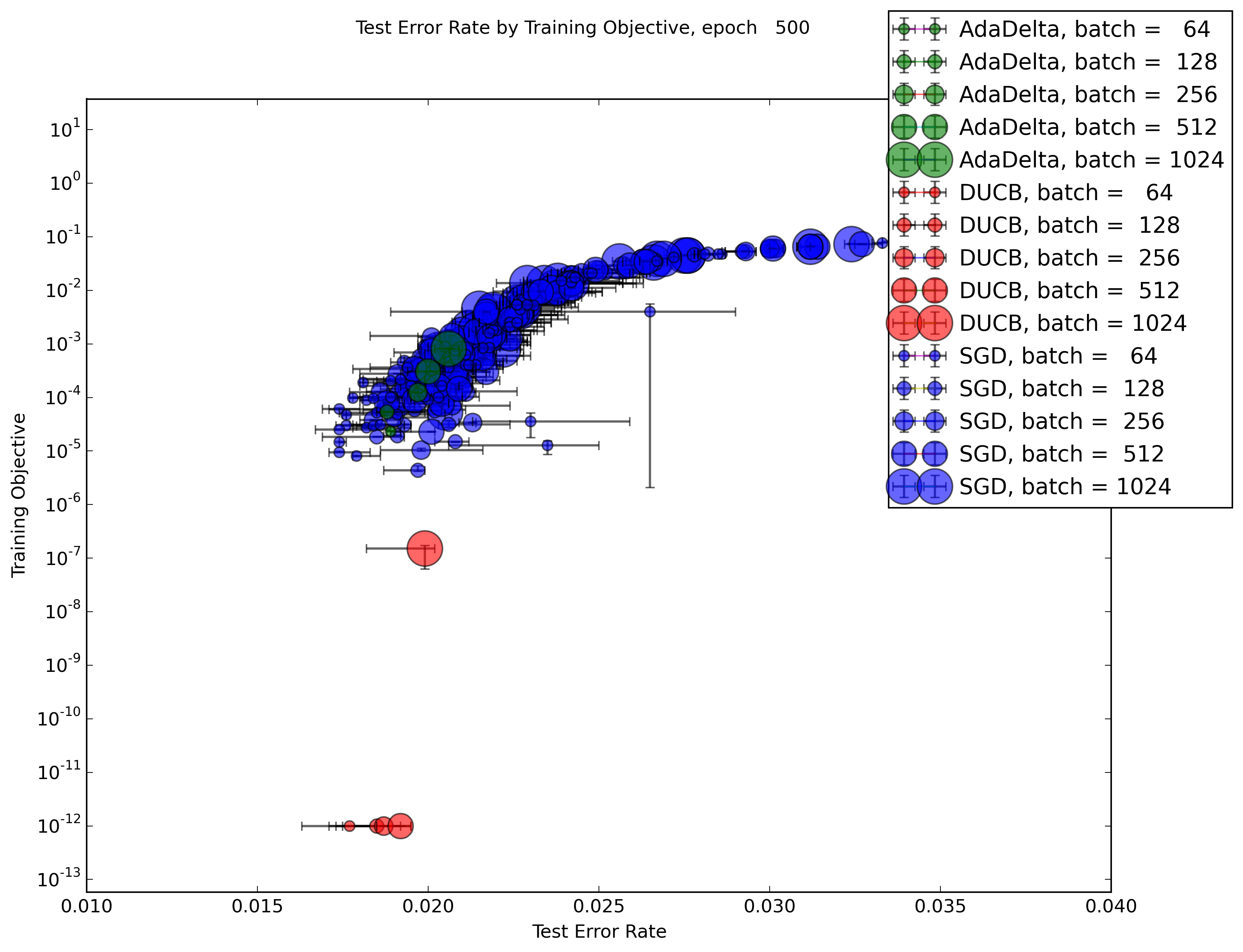}
                \caption{Epoch 500}
                \label{fig:500-300_scatter500}
        \end{subfigure}
        \caption{Best viewed in color.  Training objective (training set negative log likelihood) and test performance (test set misclassification rate) for various algorithms on MNIST 784-500-300-10 with sigmoidal units.  Each dot represents one algorithm with one set of hyperparameters. The best performing algorithms are in the bottom left.  The only parameter varied for DUCB and AdaDelta are batch size.  SGD algorithms were varied across initial learning rate, learning rate decrease schedule, momentum and batch size.  Error bars represent performance of a single algorithm and hyperparameter set across 3 random weight initializations.  All 15 instantiations of the hot swap DUCB achieve better training likelihood than all 1125 other algorithms we compare against.}
        \label{fig:500-300_scatter}
\end{figure}

We ran each algorithm for 500 epochs using three random weight initializations.  Figure \ref{fig:500-300_scatter} shows the spread of training and test performances of all 380 algorithms after 200 and 500 training epochs.  Each dot represents the median performance of a single set of hyperparameters with error bars indicating min and max performance across initializations.  The best performing algorithms will be in the bottom left.  At convergence, all 15 hot swap DUCB algorithms (red) achieve the lowest training objective (negative log likelihood) of every other algorithm tested (green and blue)

Furthermore, while test performance isn't the direct target of an optimization process---test performance is also heavily linked to regularization quality, a factor to which optimization algorithms are agnostic---the DUCB algorithm obtained the lowest median test error rate across all of its variations (1.85\% vs 1.97\% for AdaDelta and 2.35\% for SGD) and the best single test performance overall (1.63\% error rate), despite the fact that SGD instantiations outnumbered hot swapped DUCB instantiations by a factor of 75:1.

The performance of SGD algorithms vary widely across hyperparameter settings. The AdaDelta algorithms consistently exhibit performance in the top 30\% of the SGD algorithms, through as with hot swapped DUCB, they vary across batch size.

\begin{figure}[ht]
    \centering
    \includegraphics[width=\textwidth]{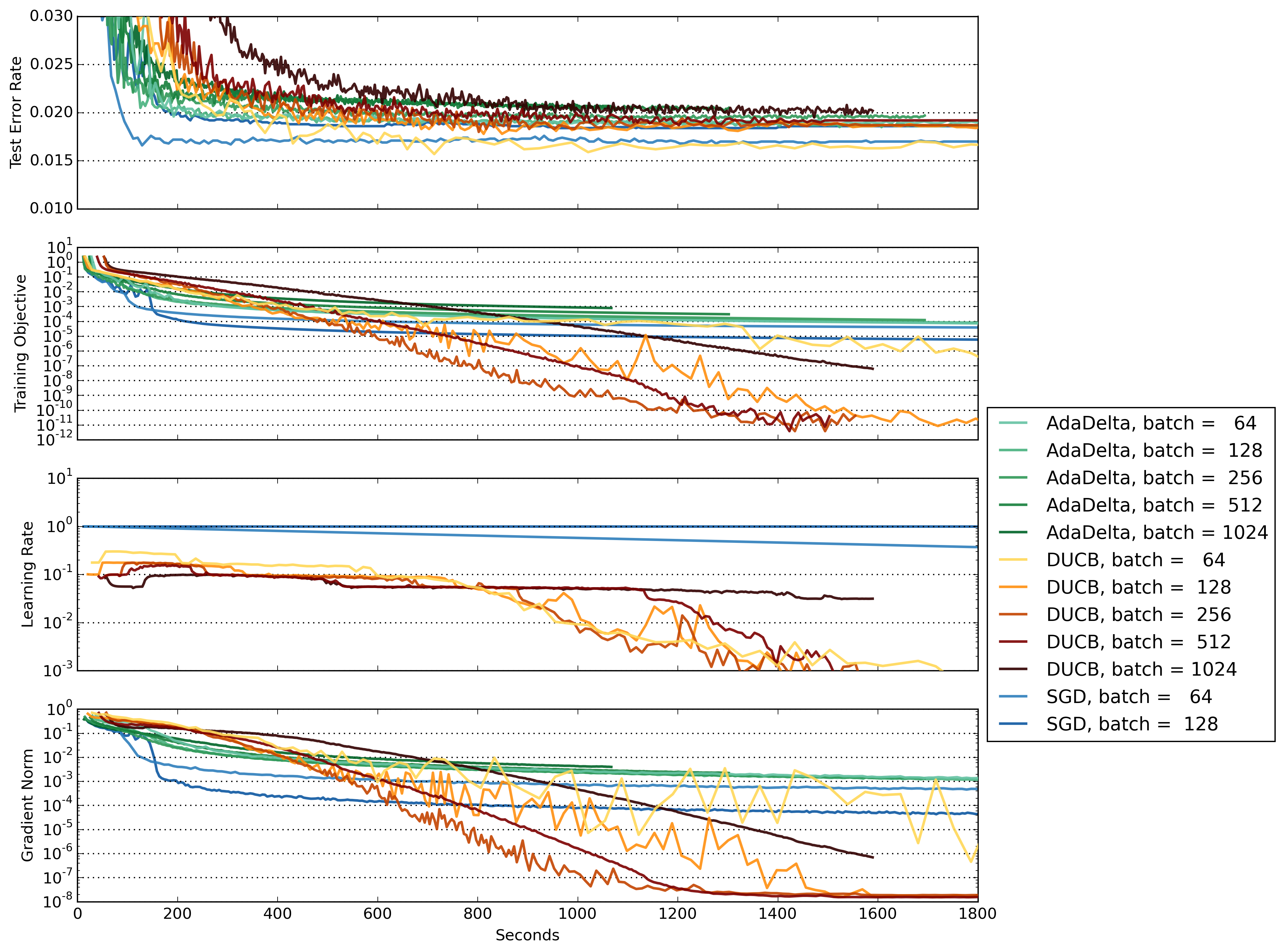}
    \caption{Best viewed in color.  Performance of hot swapping with DUCB model, AdaDelta, and the two best-performing SGD algorithms (out of a total of 1080) on MNIST 784-500-300-10 with sigmoidal units.}
    \label{fig:500-300_line}
\end{figure}

Figure \ref{fig:500-300_scatter} shows the variations in performance of different algorithms after a fixed number of training epochs, but obscures the differences in the time it takes each algorithm to complete one training epoch.  Specifically, algorithms with small batch sizes are slower because they need to perform more updates per epoch than algorithms with large batch sizes, and the DUCB algorithms are slower per iteration than the competing algorithms (see below for details).

Figure \ref{fig:500-300_line} gives a direct comparison of training time vs. various measures of performance for one instantiation each of the five DUCB hot swapping algorithms (one for each batch size), the five AdaDelta algorithms, and the two best-performing SGD algorithms (chosen retrospectivlye out of a total of 1080 SGD algorithms; `best performing' measured in final training objective and test misclassification).

The top plot in figure \ref{fig:500-300_line} shows test error rate vs. wall clock time.  Despite the fact that DUCB hot swapping with a batch size of 64 takes the most time per iteration, it makes sufficient progress in each step that by 1000 seconds into the training run it has attained the best test error of any algorithm we tested\footnote{A single iteration of DUCB hot swapping takes between 1.6 and 3.6 times as long as a comparable iteration of SGD or AdaDelta because it performs several line search iterations per step.  However, it makes considerably greater progress per iteration than SGD and continues to do so well after SGD and AdaDelta have plateaued, converging to better optima than the existing algorithms.  For additional details on the relative timing of each method, see \nameref{sec:appendix_timing}.}.  The second plot shows training objective over time (training set negative log likelihood), with all of the DUCB algorithms making significant training progress well after the SGD and AdaDelta algorithms have plateaued and finding lower optima.  The third plot suggests that the DUCB algorithms naturally learn to decrease their learning rates as they near local optima\footnote{The AdaDelta algorithms are not included in the learning rate plots because AdaDelta maintains a different learning rate for each paramter.}.  The fourth plot shows a running average of gradient norms over time, with the DUCB algorithms all finding flatter optima than competing algorithms.

On other problems in which we have tested the hot swapping DUCB procedure, we have observed similarly strong training performance with greater variation on test performance that we observed here.  This suggests that the hot swapped DUCB procedure is a strong optimization algorithm that requires similarly strong regularization not to overfit.

\section{Conclusion}
We have introduced a new adaptive learning rate algorithm for stochastic gradient that is built to hot swap an optimization hyperparameter over the course of a learning run.  Preliminary results indicate that the proposed method consistently outperforms competing methods on several measures.

Numerous extensions of this basic procedure are possible, including using different meta-models, swapping new optimization or regularization hyper parameters, swapping multiple parameters at once, and reducing the frequency of line search to speed performance.  We are currently working to test this and other hot swapping procedures on a wide variety of problems.

\subsubsection*{Acknowledgments}
This material is based upon work partially supported by  a National Science Foundation Graduate Fellowship (KB), by NSF via award number IIS-1320527 (PS), and by the Office of Naval Research under MURI grant 
N00014-08-1-1015 (PS). A portion of this work was performed by KB while a summer intern at eBay Research Labs.  

\bibliographystyle{apalike}
\bibliography{refs}

\begin{thebibliography}{}

\bibitem[Bergstra and Bengio, 2012]{bergstra_random_2012}
Bergstra, J. and Bengio, Y. (2012).
\newblock Random search for hyper-parameter optimization.
\newblock {\em The Journal of Machine Learning Research}, 13(1):281--305.

\bibitem[Gagliolo and Schmidhuber, 2006]{gagliolo2006learning}
Gagliolo, M. and Schmidhuber, J. (2006).
\newblock Learning dynamic algorithm portfolios.
\newblock {\em Annals of Mathematics and Artificial Intelligence},
  47(3-4):295--328.

\bibitem[Garivier and Moulines, 2008]{garivier_upper-confidence_2008}
Garivier, A. and Moulines, E. (2008).
\newblock On upper-confidence bound policies for non-stationary bandit
  problems.
\newblock {\em {arXiv}:0805.3415 [math, stat]}.
\newblock {arXiv}: 0805.3415.

\bibitem[Ngiam et~al., 2011]{ngiam_optimization_2011}
Ngiam, J., Coates, A., Lahiri, A., Prochnow, B., Le, Q.~V., and Ng, A.~Y.
  (2011).
\newblock On optimization methods for deep learning.
\newblock In {\em Proceedings of the 28th International Conference on Machine
  Learning ({ICML}-11)}, pages 265--272.

\bibitem[Roux et~al., 2012]{roux_stochastic_2012}
Roux, N.~L., Schmidt, M., and Bach, F.~R. (2012).
\newblock A stochastic gradient method with an exponential convergence \_rate
  for finite training sets.
\newblock In {\em Advances in Neural Information Processing Systems}, pages
  2663--2671.
\newblock 00059.

\bibitem[Schaul et~al., 2012]{schaul_no_2012}
Schaul, T., Zhang, S., and LeCun, Y. (2012).
\newblock No more pesky learning rates.
\newblock {\em {arXiv} preprint {arXiv}:1206.1106}.

\bibitem[Snoek et~al., 2012]{snoek_practical_2012}
Snoek, J., Larochelle, H., and Adams, R.~P. (2012).
\newblock Practical bayesian optimization of machine learning algorithms.
\newblock In Pereira, F., Burges, C. J.~C., Bottou, L., and Weinberger, K.~Q.,
  editors, {\em Advances in Neural Information Processing Systems 25}, pages
  2951--2959. Curran Associates, Inc.

\bibitem[Zeiler, 2012]{zeiler_adadelta:_2012}
Zeiler, M.~D. (2012).
\newblock {ADADELTA}: An adaptive learning rate method.
\newblock {\em {arXiv} preprint {arXiv}:1212.5701}.

\end{thebibliography}

\newpage
\section*{Appendix 1: Timing}
\label{sec:appendix_timing}
This section contains details of the relative timings of SGD, AdaDelta, and the hot swapped DUCB algorithm for the model detailed in section \ref{sec:results}.

DUCB tends to require more line search iterations per step as it nears a local optima, meaning that it takes less time per minibatch earlier in the optimization process and more time per minibatch later in the optimizaion process.  This effect is significantly mitigated with larger batch sizes as they exhibit less variance across minibatches.  This helps the DUCB model to predict which step size will perform optimally for a given problem, which limits the number of line search iterations required per minibatch and yields epoch timings that are closer to SGD and AdaDelta.

Overall, hot swapped DUCB takes between 1.6 and 3.6 times as long per minibatch as SGD and AdaDelta, however it makes more progress per iteration than either of these competing algorithms and converges to better optima (see figure \ref{fig:500-300_line} and the discussion in section \ref{sec:results}).

\begin{table}[ht]
  \begin{tabular}{r r r r r r}
  \hline
  \textbf{Batch Size:} & \multicolumn{5}{c}{\textbf{Milliseconds per Minibatch}} \\
  {} &  \textbf{SGD} & \textbf{AdaDelta} &  \textbf{DUCB, epoch 100} &  \textbf{DUCB, epoch 300} & \textbf{DUCB, epoch 500}  \\
  \hline
     64: &    11.9 &        12.1 &     30.1 &     40.2 &     42.4 \\
    128: &    13.0 &        13.2 &     20.3 &     38.8 &     43.7 \\
    256: &    16.8 &        16.7 &     26.4 &     39.7 &     49.2 \\
    512: &    24.7 &        25.5 &     37.9 &     38.2 &     56.2 \\
   1024: &    39.2 &        40.8 &     60.2 &     60.4 &     63.3 \\
  \hline
  \end{tabular}
  \caption{Milliseconds per minibatch for SGD, AdaDelta, and the hot swapped DUCB algorithm.  Timing varies over the course of the optimization run for the hot swapped DUCB algorithm, and so its average timing is listed after 100, 300, and 500 epochs.}
\end{table}

\begin{table}[ht]
  \begin{tabular}{r r r r r r}
  \hline
  \textbf{Batch Size:}  & \multicolumn{5}{c}{\textbf{Seconds per Epoch}} \\
  {} &  \textbf{SGD} & \textbf{AdaDelta} &  \textbf{DUCB, epoch 100} &  \textbf{DUCB, epoch 300} & \textbf{DUCB, epoch 500}  \\
  \hline
    64:  &    9.3  & 9.4 &  23.5 &  31.4 &  33.2 \\
   128:  &    5.1  & 5.2 &   7.9 &  15.2 &  17.1 \\
   256:  &    3.3  & 3.3 &   5.2 &   7.8 &   9.6 \\
   512:  &    2.4  & 2.5 &   3.7 &   3.7 &   5.5 \\ 
  1024:  &    1.9  & 2.0 &   3.0 &   3.0 &   3.1 \\
  \hline
  \end{tabular}
  \caption{Seconds per epoch for SGD, AdaDelta, and the hot swapped DUCB algorithm.  Timing varies over the course of the optimization run for the hot swapped DUCB algorithm, and so its average timing is listed after 100, 300, and 500 epochs.}
\end{table}

\newpage
\section*{Appendix 2: Full Algorithm Description}
\label{sec:appendix_algo}

\begin{algorithm}[ht]
\begin{algorithmic}[1] 
\Require $\params$, the parameters to be optimized, $\data$, a dataset which may be broken into batches denoted $\batch$, $f(\theta; \data)$ the objevtive function to be optimized,  $\Alpha$, a set of optimization hyperparameter values to consider, $M$, some model of optimization hyperparameter performance, $U(\theta; \batch, \alpha)$, an update step for the parameters $\params$ given a batch of data points and optimization hyperparameter value, a convergence criteria
	\While{not converged}
		\State $\batch \gets$ a new batch of data
		\State $\alpha \gets$ the best optimization hyperparameters $\alpha \in \Alpha$ as judged by $M$
		\State $\theta \gets U(\theta; \batch, \alpha)$   
		\State $M$ observes performance of $\alpha$
	\EndWhile
\end{algorithmic}
\caption{General hot swapped optimization}
\label{alg:hot_swap}
\end{algorithm}

\begin{algorithm}[h]
\begin{algorithmic}[1] 
\Require $\params, f, g, \alphas$ array, a dataset, convergence criteia
	\State $maxIndex \gets $ maximum index in the $\alpha$s array
	\State $rewards \gets $ array of 0s of same length as $\alphas$
	\State $counts \gets $ array of 0s of same length as $\alphas$
	\State $t \gets -1$
	\While{not converged}
		\State $t \gets t + 1$
		\State $B \gets $ new batch of data
		\State $startIndex \gets $ \Call{InitialAlphaIndex}{$rewards, counts, t, maxIndex$}		
		\State $\theta \gets $\Call{BacktrackingLineSearchWithRewards}{$f, g, \batch, \theta, \alphas, startIndex$}
	\EndWhile
\end{algorithmic}
\caption{Hot Swapped Stochastic Optimization with DUCB model}
\label{alg:hot_swap_ducb}
\end{algorithm}

\begin{algorithm}
\begin{algorithmic}[1] 
\\
\Function{InitialAlphaIndex}{$rewards, counts, t, maxIndex$}
	\If {$t < maxIndex$}
		\Return $t$
	\Else
		\Return \Call{GetDucbIndex}{$rewards, counts$}
	\EndIf
\EndFunction

\\
\Function{GetDucbSuggestedIndex}{$rewards, counts$}
	\State $rewards \gets \gamma * rewards$
	\State $counts \gets \gamma * counts$
	\State $means \gets rewards / counts$
	\State $n \gets sum(counts)$
	\State $confIntervals \gets \sqrt{exploreConst * \log(n) / counts}$
	\State $ucbs \gets means + confIntervals$
	\Return $\argmax(ucbs)$
\EndFunction

\\
\Function{BacktrackingLineSearchWithRewards}{$f, g, \batch, \params, \alphas, startIndex$}
	\State $maxIndex \gets $ the maximum index in the $\alphas$ array
	\State $f_{start} \gets f(\params; \batch)$
	\State $f_{current} \gets f_{start}$
	\State $f_{best} \gets f_{start}$
	\State $\alpha_{best} \gets startIndex$
	\State $haveFoundBetterThanStart \gets False$

	\ForAll{$index \gets startIndex : maxIndex$ }
		\State $f_{prev} \gets f_{current}$

		\State $\alpha \gets \alphas[index]$
		\State $f_{current} \gets $ \Call{ObjectiveAtAlpha}{$\params, f, g, \batch, \alpha$}
		\State \Call{GrantReward}{$index, rewards, counts, f_{start}, f_{current}$}

		\If{$f_{current} < f_{best}$}
			\State $f_{best} \gets f_{current}$
			\State $\alpha_{best} \gets \alpha$
			\State $haveFoundBetterThanStart \gets $ True
		\EndIf

		\If{$haveFoundBetterThanStart$ and $f_{current} > f_{prev}$ }
			\State Break Loop
		\EndIf
	\EndFor

	\State $\params \gets \params - \alpha_{best} g(\params; \batch)$
	\Return $\params$
\EndFunction

\\ 
\Function{ObjectiveAtAlpha}{$\params, f, g, \batch, \alpha$}
	\Return $f(\params - \alpha g(\params; \batch); \batch)$
\EndFunction

\\
\Function{GrantReward}{$index, rewards, counts, f_{start}, f_{current}$}
	\State $rewards[index] \gets rewards[index] + \log(f_{start}) - \log(f_{current})$
	\State $counts[index] \gets counts[index] + 1$
\EndFunction

\end{algorithmic}
\caption{DUCB Helper Functions}
\label{alg:ducb_helpers}
\end{algorithm}

\end{document}